\documentclass[sigconf]{acmart}

\usepackage[utf8]{inputenc} 
\usepackage[T1]{fontenc}    
\usepackage{hyperref}       
\usepackage{url}            
\usepackage{booktabs}       
\usepackage{amsfonts}       
\usepackage{nicefrac}       
\usepackage{microtype}      
\usepackage{xcolor}         
\usepackage{xspace}
\usepackage{dsfont}
\usepackage{natbib}
\usepackage{multirow}
\usepackage{bbding}
\usepackage{mathrsfs}
\usepackage{pifont}
\usepackage[american]{babel}
\usepackage{algorithm}  
\usepackage{algorithmic}  

\makeatletter
\DeclareRobustCommand\onedot{\futurelet\@let@token\@onedot}
\def\@onedot{\ifx\@let@token.\else.\null\fi\xspace}
\def\eg {\emph{e.g}\onedot} 
\def\ie{\emph{i.e}\onedot} 
 
\def\etc{\emph{etc}\onedot} 
 
\def\etal{\emph{et al}\onedot}
\makeatother

\AtBeginDocument{%
	\providecommand\BibTeX{{%
			\normalfont B\kern-0.5em{\scshape i\kern-0.25em b}\kern-0.8em\TeX}}}

\setcopyright{acmcopyright}
\copyrightyear{2022}
\acmYear{2022}
\acmDOI{XXXXXXX.XXXXXXX}

\acmConference[ASE ’22]{the 37th IEEE/ACM International Conference on Automated Software Engineering}{October 10--14,
	2022}{Ann Arbor, Michigan, USA.}
\acmPrice{15.00}
\acmISBN{978-1-4503-XXXX-X/18/06}

\begin{document}
	
	\title{
    Safety and Performance, Why not Both? Bi-Objective Optimized Model Compression toward AI Software Deployment}
	
	\author{Jie Zhu$^{1,2}$, Leye Wang$^{1,2}$, Xiao Han$^{3}$}\authornote{Corresponding author}
	\renewcommand{\authors}{Jie Zhu, Leye Wang, and Xiao Han}
	\renewcommand{\shortauthors}{Jie Zhu, Leye Wang, and Xiao Han}
	
	\affiliation{$^{1}$\institution{Key Lab of High Confidence Software Technologies (Peking University), Ministry of Education \country{China}} 
	$^{2}$\institution{School of Computer Science, Peking University, Beijing, China}
	$^{3}$\institution{Shanghai University of Finance and Economics, Shanghai, China}}
	\email{zhujie@stu.pku.edu.cn, leyewang@pku.edu.cn,  xiaohan@shufe.edu.cn}
	
	
%
	
	
	
	
	\begin{abstract}
	The size of deep learning models in artificial intelligence (AI) software is increasing rapidly, which hinders the large-scale deployment on resource-restricted devices (\eg, smartphones).
	To mitigate this issue, AI software compression plays a crucial role, which aims to compress model size while keeping high performance. However, the intrinsic defects in the big model may be inherited by the compressed one. Such defects may be easily leveraged by attackers, since the compressed models are usually deployed in a large number of devices without adequate protection. In this paper, we try to address the safe model compression problem from a safety-performance co-optimization perspective. Specifically, inspired by the test-driven development (TDD) paradigm in software engineering, we propose a test-driven sparse training framework called \textit{SafeCompress}. By simulating the attack mechanism as the safety test,  \textit{SafeCompress} can automatically compress a big model to a small one following the dynamic sparse training paradigm. Further, considering a representative attack, \ie, membership inference attack (MIA), we develop a concrete safe model compression mechanism, called \textit{MIA-SafeCompress}.  Extensive experiments are conducted to evaluate \textit{MIA-SafeCompress} on five datasets for both computer vision and natural language processing tasks. The results verify the effectiveness and generalization of our method. We also discuss how to adapt \textit{SafeCompress} to other attacks besides MIA, demonstrating the flexibility of \textit{SafeCompress}.
	\end{abstract}

	
	
	
	\maketitle
	
	\section{Introduction}
	In the last decade, artificial intelligence (AI) software, especially those based on deep neural networks (DNN), has set a huge storm~\cite{alexnet_krizhevsky2012imagenet}. Currently, AI software, with DNN as representatives, is recognized as an emerging type of software artifact (sometimes known as ``software 2.0”~\cite{remos_zhang2022remos}). Notably, the size of DNN-based AI software has increased rapidly in recent years (mostly because of the trained deep neural network model). For instance, the state-of-the-art computer vision model contains more than 15 billion parameters \cite{riquelme2021scaling}. The recent natural language model, GPT-3, is even larger, surpassing 175 billion parameters; this requires nearly 1TB of space to store only the model~\cite{brown2020language}. Such a huge model hinders realistic applications such as autonomous driving when the software requires to be deployed in resource-restricted devices like wearable devices or edge nodes. 
To this end, a new branch is derived from the traditional software compression area~\cite{Pugh1999CompressingJC,Ernst1997CodeC,drinic2007ppmexe,Wolfe1992ExecutingCP}, called \textit{AI software compression} (especially DNN \textit{model compression}\footnote{In the rest of the paper, without incurring the ambiguity, we will use `model compression' to indicate the `DNN model compression in AI software' for clarity.}), and has attracted a lot of research interest to date.
	
	
	Model compression aims to compress a large DNN model to a smaller one given specific requirements \eg, parameter numbers, model sparsity, and compression rate. Rashly compressing a model may lead to severe degeneration in the AI software's task performance such as classification accuracy. To balance memory storage and task performance, many compression methods have been proposed and deployed~\cite{pruning_han2015deep, kd_ba2014deep, KD_hinton2015distilling, Tiny_BERT_jiao2019tinybert, quantify_chen2015compressing, quantify_han2015learning}. 
	For example, Han \etal~\cite{quantify_han2015learning} prune AlexNet~\cite{alexnet_krizhevsky2012imagenet} and reduce its size  by 9 times while losing only 0.01\% accuracy in image classification. Jiao \etal~\cite{Tiny_BERT_jiao2019tinybert} reduce the size of BERT~\cite{devlin2018bert} by about 2 times via knowledge distillation while losing a 0.1\%  average score in the GLUE~\cite{wang2018glue} official benchmark.
	
	
	While the compressed model aims to mimic the original model's behavior, its defects may also be inherited. As a representative case, big deep models are verified to be able to memorize training data~\cite{secret_carlini2019secret}, thus leading to private data leakage when facing threats such as membership inference attacks~\cite{Shaow_Learning}; such a vulnerability would probably remain in the compressed model. More seriously, model compression is often used for AI software deployment on a large number of edge devices (smartphones, wearable devices, \etc)~\cite{Deng2020ModelCA, liu2020ensemble}; compared to the big model (often stored in a well-maintained server), attackers thereby have much more opportunities to access to compressed models and attack them
	 (\eg, an attacker may act as a normal user to download the compressed model in her/his own device). \textit{In a word, compressed models will inherit the vulnerabilities of big models, while facing even higher risks of being attacked~\footnote{Besides accessing to compressed models more easily for attackers, we also find that some compressed models have higher attack accuracy in our experiments, which means these compressed models are more vulnerable.}}. Hence, studying how to do \textbf{safe model compression} is urgently required.
	
	An intuitive solution to the safe model compression is directly combining two streams of techniques, forming a \textit{two-step solution}: (i) model compression and (ii) model protection. For instance, we can first obtain a small model using existing compression techniques, and then apply protection techniques (\eg, differential privacy~\cite{dp_abadi2016} and knowledge distillation~\cite{kd_membership}) to improve the model safety against certain attacks. \textit{However, the two-step solution may suffer from poor model performance~\footnote{We use `performance' as a general term to describe any specific task-dependent metric. For instance, if a model is developed for image classification, the model performance can be measured by the classification accuracy metric.} and low safety due to lack of consideration of the interaction between the two techniques.} For example, Yuan \& Zhang~\cite{pruning_defeat_yuan2022membership} find that pruning makes the divergence of prediction confidence and sensitivity increase and vary widely among different classes. This may not be sensed by defenders for lack of interaction but can be manipulated by attackers.
	
	In this paper, we try to address the safe model compression problem from a \textit{performance-safety co-optimization} perspective. More specifically, inspired by \textit{test-driven development}~\cite{Beck2003TestdrivenD} and \textit{dynamic sparse training}~\cite{DST_Begin_mocanu2018scalable}, we propose a \textit{test-driven sparse training framework for safe model compression}, called \textbf{\textit{SafeCompress}}. By simulating the attack mechanism to defend, \textit{SafeCompress} can automatically compress a big model into a small one (with required sparsity) to optimize both model performance and safety. \textit{SafeCompress} generally follows an iterative optimization approach. For initialization, \textit{SafeCompress} randomly prunes a big model to a sparse one, which serves as the input of the first iteration. In each iteration, various compression strategies are applied to the input model to derive more sparse models; then, \textit{SafeCompress} launches a performance-safety co-optimization mechanism that applies both task performance and simulated-attack-based safety tests on the derived sparse models to select the best compression strategy. Then, the sparse model with the selected (best) strategy becomes the input of the next iteration. The iterative process will terminate after a predefined maximum number of iterations or a new model has little improvement in performance and safety tests.

	Based on the \textit{SafeCompress} framework, we design and implement a concrete safe model compression mechanism against membership inference attacks (MIAs)~\cite{Shaow_Learning}, denoted as \textit{\textbf{MIA-SafeCompress}}. In \textit{MIA-SafeCompress}, we further propose an entropy-based regularizer that increases the uncertainty of model outputs to protect the model from MIAs. Note that we choose MIA as the attack example because MIA is a representative method for evaluating AI model safety, especially from the privacy leakage aspect~\cite{dp_yeom2018privacy,nasr2019comprehensive}.

	Our contribution can be summarized as follows:
	
	$\bullet$ To the best of our knowledge, this is one of the pioneering efforts toward the \textit{safe model compression} problem, which is critical for today's large-scale AI software deployment on edge devices such as smartphones.
	
	$\bullet$ To address the safe model compression problem, we propose a general framework called \textit{SafeCompress}, which can be configured to protect against a pre-specified attack mechanism. In brief, \textit{SafeCompress} adopts a test-driven process to iteratively update the model compression strategies to co-optimize model performance and safety. 
	
		
		$\bullet$ Considering MIA as a representative attack mechanism~\cite{ResAdv}, we develop a concrete instance of \textit{SafeCompress}, \ie, \textit{MIA-SafeCompress}. We further enhance the iterative model compression process with a new entropy-based regularizer to increase the model safety against MIA. We also discuss how to adapt \textit{SafeCompress} to other attacks.
		
		$\bullet$ Using \textit{MIA-SafeCompress} as a showcase of \textit{SafeCompress}, we conduct extensive experiments on five datasets of two domains (three computer vision tasks and two natural language processing tasks). Results verify that our method significantly outperforms baseline solutions that integrate state-of-the-art compression and MIA defense techniques. The code of \textit{MIA-SafeCompress} is available as open source at \url{https://github.com/JiePKU/MIA-SafeCompress}.

	\section{Background}

	\subsection{Test-Driven Development}
    Test-driven development (TDD)~\cite{Beck2003TestdrivenD} is a programming paradigm where test codes play a vital role during the whole software development process. With TDD, before writing the codes for the software functionality, programmers would write the corresponding test suite in advance; then, the test suite can justify whether the software functionality is implemented properly or not. In general, TDD leads to an \textit{iterative coding-testing} process to improve the correctness and robustness of the software; it has become a widely-adopted practice in software development (\eg, SciPy~\cite{Virtanen2020SciPy1F}). Inspired by TDD, we develop a test-driven safe model compression framework called \textit{SafeCompress}. Similar to the iterative coding-testing process in TDD, \textit{SafeCompress} adopts an \textit{iterative compressing-testing} process. In particular, by specifying the attacks to fight, \textit{SafeCompress} can automatically update the compression strategies step by step to optimize model performance and safety simultaneously.

	\subsection{Dynamic Sparse Training}
	Dynamic sparse training (DST) is a sparse-to-sparse training paradigm to learn a sparse (small) DNN model based on a dense (big) one~\cite{DST_Begin_mocanu2018scalable,mocanu2017network}. Specifically, it starts training with a sparse model structure initialized  from a dense model.  As training progresses, it modifies the architecture iteratively by pruning some neural network connections and growing new connections based on certain strategies. This enables neural networks to explore self-structure until finding the most suitable one for the training data. 
	Note that \textit{SafeCompress}'s iterative updating strategy for optimizing the compressed model structure is just inspired by the DST paradigm. 
	
	\subsection{Membership Inference Attacks}
	Prior research has extensively verified 
	that DNN models often exhibit different behaviors on training data records (\ie, members) versus test data records (\ie, non-members); the main reason is that models can memorize training data during the repeated training process lasting for a large number of epochs~\cite{secret_carlini2019secret,nasr2019comprehensive}. For instance, a DNN model would generally give a higher prediction confidence score to a training data record than a test one, as the model may remember training data's labels. Based on such observations, membership inference attacks (MIAs)~\cite{Shaow_Learning} are proposed to build attack models to infer whether one data record belongs to training data or not. When it comes to sensitive data,  personal privacy is exposed to a great risk. For example, if MIA learns that a target user's electronic health record (EHR) data is used to train a model related to a specific disease (\eg, to predict the length of stay in ICU~\cite{ma2021distilling}), then the attacker knows that the target user has the disease.
	In this paper, we use MIA as an attack instance to verify the feasibility and effectiveness of \textit{SafeCompress}, because MIA has become one of the representative attack methods to evaluate the safety of DNN models both theoretically and empirically~\cite{dp_yeom2018privacy, nasr2019comprehensive}.

	\section{Problem Formulation}

    Given a big model $\mathcal{F}(;\theta)$ parameterized by $\theta$, we aim to find a sparse model $\mathcal{F}(;\hat \theta)$ (most elements in $\hat \theta$ are zero) under certain memory restriction $\Omega$ and the sparse model can defend against a pre-specified attack mechanism $f_A$. We use $G_{f_A}$ to denote the attack gain of $f_A$. We restrict the compression ratio, or called model sparsity, below $\Omega$ (\ie, the percentage of non-zero parameters in the sparse model over the original model).
	We aim to minimize both the task performance loss $\mathcal L$ and the attack gain $G_{f_A}$ over the sparse model $\mathcal{F}(;\hat \theta)$:
	\begin{align} 
	\label{general_problem_L}
	& \min_{\hat \theta} \sum_ {x,y}  \mathcal{L} (\mathcal{F}(x;\hat \theta), y) \\
    \label{general_problem_G}
	& \min_{\hat \theta} G_{f_A} (\mathcal{F}(;\hat \theta)) \\
    \label{general_problem_S}
	& \operatorname{ s.t. }  \frac{{\; \lVert \hat \theta \rVert}_{0}}{{\lVert \theta \rVert}} \leq \Omega,
	\end{align}
	where $x$ is a sample and $y$ is the corresponding label, and $\mathcal{L}$ represents a
	task-dependent loss function. ${\lVert \cdot  \lVert}_{0}$ counts the number of non-zero elements, and ${\lVert \cdot  \lVert}$ calculates the number of all the elements. Note that it is a bi-objective optimization problem regarding both model performance (Eq.~\ref{general_problem_L}) and safety (Eq.~\ref{general_problem_G}).
	
	The above formulation is a general one without specifying the attack mechanism $f_A$. In this research, we use MIA as an example of $f_A$. The gain $G_{f_A}$ for MIA is then formulated as follows.
	
	\textbf{MIA gain}. MIA infers whether a sample $x$ is in the training dataset or not. To this end, MIA usually trains a binary classification model $f_A$. In this work, we consider $f_A$ as a neural network classifier with the black-box setting~\cite{Shaow_Learning} (\ie, only model outputs are available to $f_{A}$).\footnote{Other MIA settings, \eg, white-box~\cite{nasr2019comprehensive}, can also be supported by our \textit{SafeCompress} framework. Sec.~\ref{sub:configure_other_attacks} includes a detailed discussion.} Then, given training samples (${D}_{tr}$) and non-training samples (${D}_{\
\neg tr}$), the expected gain of $f_A$ is:
	\begin{equation} \label{G_}
		\begin{split}
		G_{f_A}(\mathcal F(;\hat \theta))= & \sum_ {x, y}  \mathds{1}(x \in D_{tr})\log (f_A(\mathcal F(x;\hat \theta),y)) +  \\
				& \mathds{1}(x \in D_{\neg tr})\log (1-(f_A(\mathcal F(x;\hat \theta),y)))  ,
		\end{split}
	\end{equation}
	where $\mathds{1}(x \in Z)$ is 1 if the sample $x$ belongs to $Z$, otherwise 0. 

	\section{Method} 
	
	\subsection{Key Design Principles}
	
	We clarify two key principles driving our design, \ie, \textit{attack configurability} and \textit{task adaptability} before elaborating on design details.
	    
	\textbf{Attack Configurability}. In reality, adversaries may conduct various types of attacks \cite{han2019f}. Hence, a practical solution should be able to do safe compression against an arbitrary pre-specified attack mechanism. In other words, the proposed solution should be easily configured to fight a given attack mechanism.
	
	\textbf{Task Adaptability}. AI (especially deep learning) techniques have been applied to various task domains including computer vision (CV), natural language processing (NLP), \etc To this end, a useful solution is desired to be able to adapt to heterogeneous AI tasks (\eg, CV or NLP) very easily.

    Our design of \textit{SafeCompress} just follows these two principles, thus ensuring practicality in various AI software deployment scenarios. Next, we describe the design details.
    
	\subsection{SafeCompress: A General Framework for Safe Model Compression} \label{defense}
\begin{figure*}[t]
		\centering{\includegraphics[width=.9\linewidth]{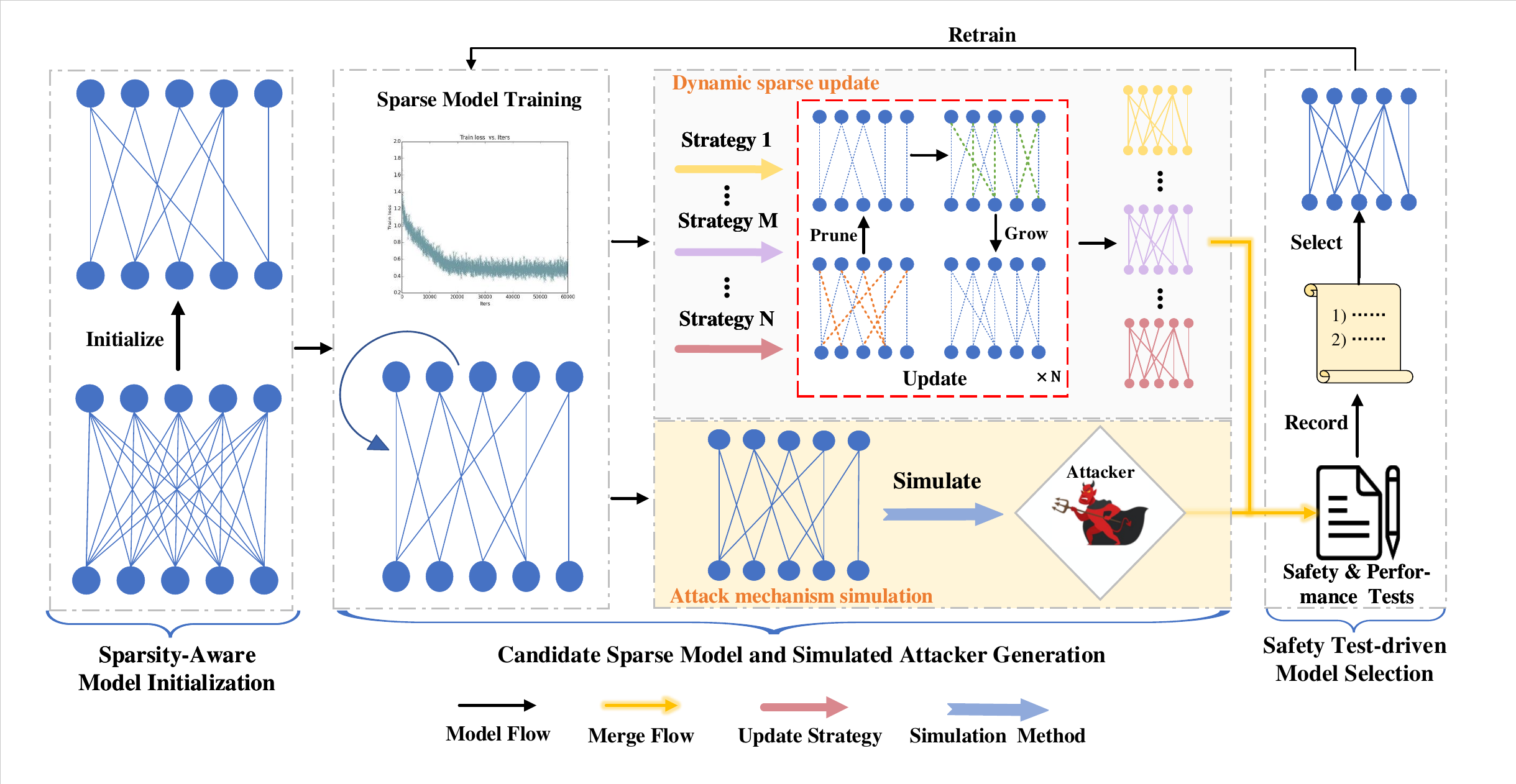}}
		\vspace{-1em}
		\caption{An overview of our framework SafeCompress.}
		\label{overview}
	\end{figure*}

	As shown in Figure~\ref{overview}, \textit{SafeCompress} contains three stages.
	
        \textbf{Stage 1. Sparsity-Aware Model Initialization.} $\:$ The sparsity (\ie, compression ratio) needs to be considered as a priority. 
		We follow the dynamic sparse training paradigm to restrict the sparsity in model initialization. Specifically, based on a given big model (an arbitrary deep model for various tasks like CV or NLP), we initialize a sparse one to meet memory requirements. After initialization, the sparse model is sent to Stage 2.
		
		\textbf{Stage 2. Candidate Sparse Model and Simulated Attacker Generation.} $\;$ During this stage, the sparse model is firstly trained until reaching the stopping criteria. Then, the trained model is fed to two branches. The first branch is called \textit{dynamic sparse update} where different combinations of pruning and growth strategies are performed on the input model, producing new candidate model variants with diverse sparse architectures. Note that the number of removed connections equals that of the reactivated ones, guaranteeing the sparsity unchangeable. The second branch is the \textit{attack mechanism simulation}. In this branch, we simulate an external attacker that aims to attack candidate sparse models. The simulated attacker and candidate sparse models are sent to Stage 3.  
		
		\textbf{Stage 3. Safety Test-driven Model Selection.} $\;$ A safety test is performed on these input sparse models by the simulated attacker. The one that performs the best in this test will be selected and sent back to Stage 2, starting a new iteration. 
		The whole process will terminate after running for a predefined number of iterations.
    \begin{algorithm}[t]  
	\small
		\caption{SafeCompress Framework Procedure}  
		\label{alg:Framwork}  
		\begin{algorithmic}[1]  
			\REQUIRE A big model $\mathcal{M}_{L}$; A sparsity requirement $\Omega$; A set of update strategies $\mathcal{U}$ and the size is $N$; Training stopping criteria for a sparse model $\mathcal{T}$; Total epochs for termination $Eps$; 
			\ENSURE  
			A well-trained sparse model $\mathcal{M}_{S}$;
			\STATE Initialize $\mathcal{M}_{L}$ as a sparse one to meet the sparsity requirement $\Omega$; We denote this sparse model as $\mathcal{M}_{S}$
			\STATE Train $\mathcal{M}_{S}$ until  condition $\mathcal{T}$ is satisfied \textbf{then do}: \label{code:fram:extract}  
			\STATE \textbf{for} each $\mathcal{U}_{i}$ in $\mathcal{U}$ \textbf{do}:
			\STATE \quad Update $\mathcal{M}_{S}$ via $\mathcal{U}_{i}$ denoted as $\mathcal{M}_{S}^{i}$ 
			\STATE  \textbf{end for}	
			\STATE  Obtain a candidate sparse model set $\mathcal{C}$=\{$\mathcal{M}_{S}^{1}$...$\mathcal{M}_{S}^{N}$\}
			\STATE  Simulate an external attacker as $\mathcal{A}$ (with the help of $\mathcal{M}_{S}$)
			
			\STATE		$\mathcal{M}_{S}^{best}$ $\leftarrow$ Pick the best from \{safety test($\mathcal{C}$, $\mathcal{A}$)\}
			\STATE $\mathcal{M}_{S}$ $\leftarrow$	$\mathcal{M}_{S}^{best}$
			\STATE \textbf{if not} achieve total epochs $Eps$:  	   			  
			\STATE \quad  \textbf{go to} \ref{code:fram:extract}
			\quad \RETURN $\mathcal{M}_{S}$  
		\end{algorithmic}  
		\vspace{-.4em}
	\end{algorithm} 
	
    The pseudo-code of \textit{SafeCompress} is in Algorithm~\ref{alg:Framwork}. Note that we do not restrict the type of attacks in \textit{SafeCompress}. It thus has a potential to prevent various attacks toward AI software and models. Next, we would instantiate the \textit{SafeCompress} framework for MIA. Afterward, we discuss how to adapt \textit{SafeCompress} to other attacks.
	
	
	\subsection{MIA-SafeCompress: Defending Membership Inference Attacks based on SafeCompress} \label{implementation}
	We implement a concrete safe model compression mechanism against MIA, called \textit{MIA-SafeCompress} based on \textit{SafeCompress}.
	
	\subsubsection{Stage 1. Sparsity-Aware Model Initialization}
	
	Given a big model,  we adopt the Erdös–Rényi~\cite{initialization_method,chen2021chasing} initialization method to reach a predefined sparsity requirement, \ie, removing a number of model connections (assigning zeros to the connection weights). Specifically, for the $k$-th layer with $n_k$ neurons, we collect them in a vector and denote it as $V^k = [v^{k}_{1}, v^{k}_{2}, v^{k}_{3},...... v^{k}_{n_k}]$. Usually, $V^k$ in the $k$-th layer is connected with last layer $V^{k-1}$ via a weight matrix $W^k \in \mathbb{R}^{n_k \times n_{k-1}}$. In the sparse setting, the matrix degenerates to a Erdös–Rényi random graph, where the probability of connection between neuron $V^{k}$ and $V^{k-1}$ is decided by:
	\begin{equation}
		P(W_{ij}) = \frac{\epsilon \times (n_{k} + n_{k-1} )}{n_{k} \times n_{k-1}} \,,
	\end{equation} 
	where $\epsilon$ is a coefficient that is adjusted to meet a target sparsity.
	In general, this distribution is inclined to allocate higher sparsity (more zeros) to the layers with more parameters as the probability tends to decline while parameters increase.\footnote{For the weights of remained connections, we can keep those of the big model, or simply do random initialization. In our experiments, we find that keeping big model weights performs not better than random initialization. Hence, we adopt random initialization in our implementation.}
	
	\subsubsection{Stage 2. Candidate Sparse Model and Simulated Attacker Generation} 
	
	During this stage, we train the input sparse model for a predefined number of iterations (following the setting in~\cite{DST_InTime}, the iteration number is set to 4,000). Once finished, the well-trained sparse model, denoted as $\mathcal{M}_{S}$, is fed into two branches. 
	
	\textbf{Branch 1. Dynamic sparse update}. In the first branch, \ie, the dynamic sparse update branch, we apply two state-of-the-art pruning strategies and two growth strategies to operate on $\mathcal{M}_{S}$, leading to four ($2*2$) different sparse typologies. The two pruning strategies are \textit{magnitude-based} pruning~\cite{pruning_han2015deep, Dai2019Magnitude} and \textit{threshold-based} pruning~\cite{quantify_han2015learning}. Magnitude-based pruning removes the connections with the smallest weight magnitudes; threshold-based pruning removes the connections whose weight magnitudes are below a given threshold. The two growing strategies are \textit{gradient-based} growth~\cite{evci2020rigging,weight_redistribut_dettmers2019sparse} and \textit{random-based} growth~\cite{DST_Begin_mocanu2018scalable}. The gradient-based growth reactivates connection (weight) that has a large gradient $\lvert \frac{\partial \mathcal{L}}{\partial w}\rvert$,
	while the random-based growth randomly reactivates connection. Afterward, we fine-tune these derived sparse models and generate four candidate sparse models $\mathcal M_{S}^1,...,\mathcal M_{S}^4$.\footnote{As more advance pruning or growth strategies may be proposed in the future, \textit{SafeCompress} is easy to incorporate them by adding to/replacing existing strategies.}
	
\begin{figure} 
		\centering{\includegraphics[width=.9\linewidth]{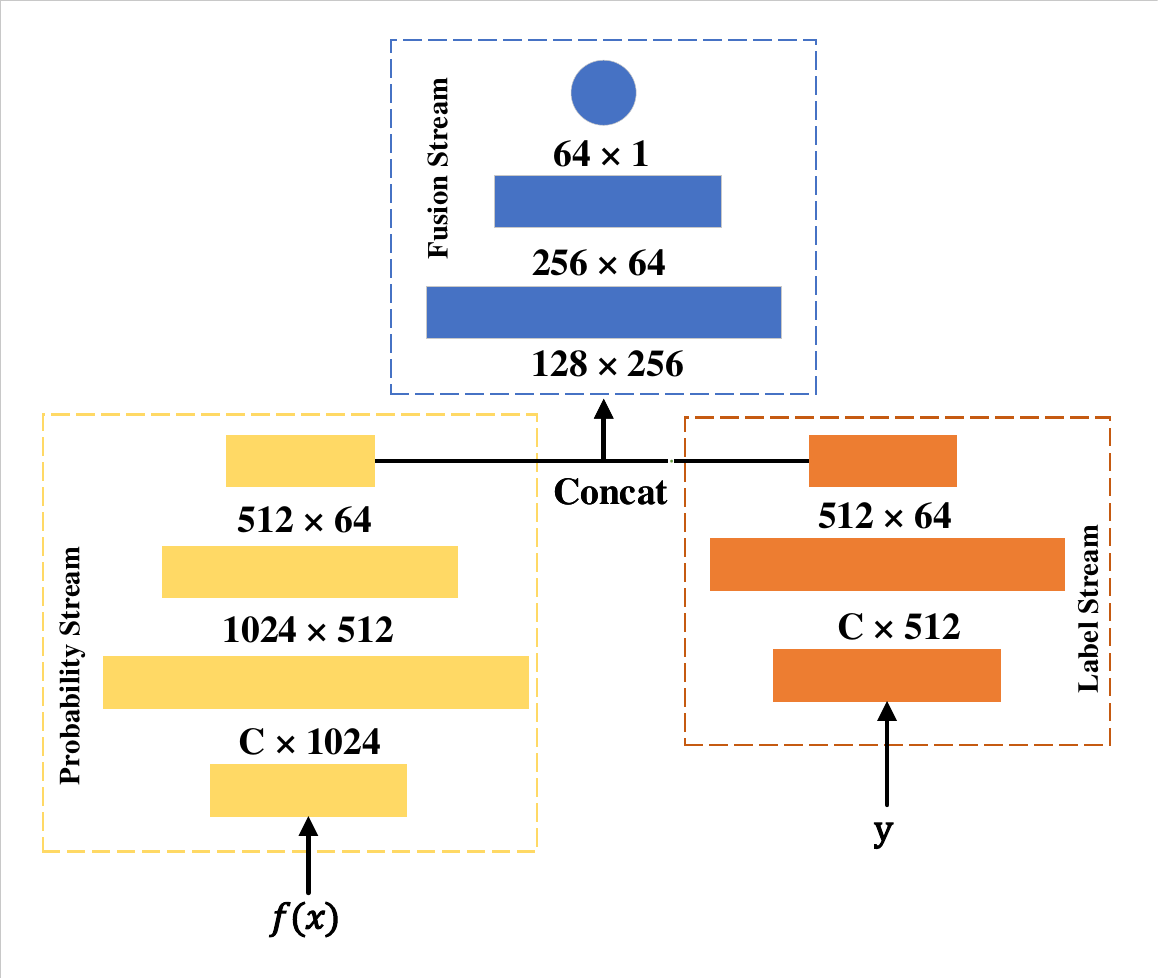}}
	    \vspace{-1em}
		\caption{The architecture of the simulated MIA neural network model. 
		Each layer is fully connected to its subsequent layer. $C$ is the category number.}
		\label{attack}
		\vspace{-1em}
	\end{figure}

	\textbf{Branch 2. Attack mechanism simulation}. We try to simulate an external attacker in preparation for safety tests in the second branch. Specifically, we follow the previous MIA work~\cite{ResAdv} and simulate the MIA attacker with a fully connected neural network, as depicted in Figure~\ref{attack}. The simulated attacker contains three parts: \textit{probability stream}, \textit{label stream}, and \textit{fusion stream}. The probability stream processes the output probability from the target sparse model $\mathcal{M}_{S}(x)$. The label stream deals with the label $y$ of the sample $x$. Then, the fusion stream fuses the features extracted from the above two streams and outputs a probability to indicate whether the sample is used in training or not. Note that to improve the attacker simulation efficiency, we do not independently train an attacker for each candidate sparse model $\mathcal M_s^i$ in Branch 1. Instead, we first train a simulated attacker $\mathcal A$ based on $\mathcal M_S$; then, for each $\mathcal M_s^i$, we fine-tune $\mathcal A$ for several epochs to ensure its attack effectiveness toward $\mathcal M_s^i$. This training acceleration strategy is sensible as $\mathcal M_s^i$ often deviates from $\mathcal M_s$ only marginally.
	
	Finally, four sparse model variants (Branch 1) and  corresponding well-trained simulated attackers (Branch 2) are sent to the next stage for safety tests.
	
	\subsubsection{Stage 3. Safety Test-driven Model Selection.} 
	
	Safety test-driven model selection aims to choose the candidate sparse model with the best trade-off in task performance and safety protection. Specifically, we employ the simulated attacker to conduct MIA safety tests on four candidate sparse models and then record the attack accuracy. Subsequently, we can simply select the one whose attack accuracy is the lowest (strong defense ability). However, this manner ignores the task performance, thereby missing a comprehensive consideration. Hence, we also conduct task performance tests on four sparse models  (\eg, image classification accuracy for CV tasks). Then, by considering the attack accuracy (safety) and task performance together, we can select the best candidate model. In our implementation, we use a newly-defined \textbf{TM-score} (\ie, \textbf{T}ask performance divided by \textbf{M}IA attack accuracy, details in Sec.~\ref{sub:attack_setup}) for model selection. The candidate sparse model with the highest TM-score is selected. 
	The selected model is then sent back to Stage 2 and a new iteration starts. The whole process will terminate after repeating a predefined number of iterations.
	
	\subsubsection{Metric}\label{sub:attack_setup} For performance, we evaluate model accuracy on classification tasks, represented by \textit{Task Acc}. The higher the Task Acc is, the better the model performs. For safety, we adopt the membership inference attack accuracy, denoted as \textit{MIA Acc}, to reflect defense ability. The lower the MIA Acc is, the stronger the model's defense ability is. Besides, we take both performance and safety into consideration, and design a metric called \textit{TM-score} that aims to directly evaluate the performance-safety trade-off:
    \begin{equation}\label{score}
    TM\textbf{-}score = \frac{{\quad(Task\; Acc)}^{\lambda}}{\;MIA\; Acc} \,,
    \end{equation}
    where $\lambda$ is a coefficient to control this trade-off and we set $\lambda=1$ in our method for simplicity. 
    If a model keeps high Task Acc and low MIA Acc, the resulted TM-score will be high. 
    In brief, this metric is in line with our goal to seek a model with both good performance and strong safety.
	
	\subsubsection{Enhancement with Entropy-based Regularization} \label{re}

	MIA mechanisms usually perform inference attac ks by taking the target model's output probabilities as inputs. Consequently, we consider to conceal the discriminative information by increasing the uncertainty in output probability. In information theory, entropy has been employed to describe the degree of uncertainty. For a \textbf{\textit{c}}-class classification task, given its output probability distribution, the entropy is formulated as:
	\begin{equation}
		E = - \sum_{i=0}^{c-1}  p_i \log (p_i)
	\end{equation} 
	It is easy to know that $E$ gets the maximum value when $p_i$ is equal to $\frac{1}{c}$. Intuitively, it is hard for a classifier to distinguish the right category if the output possibility distribution is not biased. In the worst case when each output class has an equal probability, it degenerates to a random guess. Fortunately, this behavior meets our expectations when defending against MIA. In other words, increasing entropy could enhance the target model's ability to defend against MIA. Hence, we define a new task model training loss as follows: 
	\begin{equation}
		\mathcal{L} = \mathcal{L}_r  -  \beta \cdot \frac{1}{N}\sum_{j=0}^{N-1} E_j \,,
	\end{equation} 
	where $\mathcal{L}_r$ is a classification loss (\eg, cross-entropy), $\beta$ is a coefficient to weight the entropy term, and $N$ is the batch size. We name this loss \textit{\textbf{re1}}. We set $\beta$ to $0.1$ and $N$ to $128$ in our implementation.
	
	Further, we propose another loss variant called \textbf{\textit{re2}}. 
	Inspired by previous works~\cite{raw_re_pereyra2017regularizing, re2_larrazabal2021maximum}, we design a fine-grained regularization term that considers the classification result of each sample, while \textit{re1} just operates on every sample no matter whether it is classified correctly.  Specifically, 
	we denote $\Phi$ as a set where misclassified samples are collected in a batch $N$.  Thus, \textit{re2} is formulated as:
	\begin{equation}
		\mathcal{L} = \mathcal{L}_r  -  \beta \cdot \frac{1}{\Vert \Phi \Vert }\sum_{j \in \Phi} E_j \,,
	\end{equation} 
	where $\Vert \Phi \Vert$ is the size of set $\Phi$. We keep the same setting as \textit{re1} for $\beta$ and $N$.

	\subsection{Configuring SafeCompress to Other Attacks}
	\label{sub:configure_other_attacks}
	
	To adapt \textit{MIA-SafeCompress} to other attacks, the main modification is using another attack mechanism instead of MIA in Branch 2 of Stage 2. Some examples are listed as follows.
	
	\textbf{White-box MIA}. Different from our implemented black-box MIA, white-box MIA assumes that attackers have extra knowledge about the target model beyond outputs (\eg, hidden layer parameters~\cite{sablayrolles2019white} and gradient descents in training epochs~\cite{nasr2019comprehensive}). Then, to implement a white-box MIA mechanism, we can just include the extra information as inputs to simulate the attacker mechanism $\mathcal A$. In \textit{Appendix}, we have conducted a preliminary experiment to verify the feasibility of extending our framework to white-box MIA.
	
	\textbf{Attribute Inference Attack}. Attribute inference attack (AIA) aims to infer private information of a user (sample), such as age and location~\cite{han2019f}. Suppose that the original big model (\eg, auto-encoder~\cite{bengio2013representation}) is trained to obtain a representation from a user's movie ratings to serve downstream tasks like recommendation; this representation may leak the user's private information~\cite{kosinski2013private}. Then, we can simulate the attacker $\mathcal A$ as a specific AIA method (\eg, logistic regression~\cite{kosinski2013private}), or a basket of multiple AIA methods~\cite{han2019f}. Note that the safety and performance metrics also need to be modified. For safety, the metric depends on the private attribute --- \eg, to protect age inference, the metric could be set to mean absolute error. For performance, we can list several downstream tasks and then evaluate the learned representations' effects on these tasks.
	
 	\section{Experimental Setup}
 	\subsection{Datasets and Models}
 	We conduct experiments on five datasets. For each dataset, we select one model widely used for the dataset task as the original big model. The training/test data partition setting follows the reference papers.
 	
 	\textbf{CIFAR10 and CIFAR100}~\cite{cifar_10_krizhevsky2009learning} are two benchmark datasets for image classification. Both of them have $50,000$ training images and $10,000$ test images. CIFAR10 has $10$ categories while CIFAR100 has $100$ categories. The size of every image is $32 \times 32$. We adopt AlexNet~\cite{alexnet_krizhevsky2012imagenet} for CIFAR10 and VGG16~\cite{vgg_simonyan2014very} for CIFAR100.
 	
 	\textbf{Tinyimagenet}~\cite{tinyimagenet_le2015tiny} is another image dataset that contains $200$ categories. Each category includes $500$ training images and $50$ test images. The size of each image is $64 \times 64$. We adopt ResNet18~\cite{resnet_he2016deep} for Tinyimagenet.
 	
 	\textbf{Yelp-5}~\cite{Yelpzhang2015character} is a  review dataset for  sentiment classification (5 categories). It includes $130,000$ training and $10,000$ test texts per class. We adopt BERT~\cite{BERT_devlin2018bert} for Yelp-5.
 	
 	\textbf{AG-News}~\cite{AG-Newszhang2015character} is a topic classification dataset. It has 4 classes. For each class, it contains 30,000 training and 1,900 test texts. We adopt Roberta~\cite{roberta_liu2019roberta} for AG News.
 	   
    \textbf{Dataset Splits for Attacker Simulation.} We split experimental datasets as shown in Table~\ref{data_split} to evaluate the model defense ability.  Following previous works~\cite{Pruning_IJCAI, kd_membership, MIAs_KD_V2}, we assume the simulated attacker knows 50\% of the (target) model's training data and 50\% of the test data (non-training data), which are used for training the attack model. The remaining datasets are adopted for evaluation.
 	\begin{table}
    \small
	\caption{Number of samples in dataset splits.}
	\label{data_split}
	\vspace{-1em}
	\begin{tabular}{l c  c | c  c}
		\toprule
		\multirow{2}{*}{\textbf{Datasets}}&
		\multicolumn{2}{c}{\textbf{Attack Training}}&\multicolumn{2}{c}{\textbf{Attack Evaluation}}\cr
		\cmidrule(lr){2-3} \cmidrule(lr){4-5}
		& $D^{known}_{train}$& $D^{known}_{test}$ &
		$D^{unknown}_{train}$ & $D^{unknown}_{test}$\\
		\midrule
		\textit{CIFAR10} & 25,000 & 5,000 & 25,000 & 5,000 \\
		\textit{CIFAR100} & 25,000 & 5,000 & 25,000 & 5,000 \\
		\textit{Tinyimagenet} & 50,000 & 5,000 & 50,000 & 5,000 \\ 
		\midrule
		\textit{Yelp-5} & 325,000 & 25,000 & 325,000 & 25,000 \\ 
		\textit{AG-News} & 60,000 & 3,800 & 60,000 & 3,800 \\ 
		\bottomrule
	\end{tabular}
\end{table}
 	 \setlength{\tabcolsep}{0.2cm}\begin{table}
 	    \small
 		\caption{Two-step baselines.}
 		\label{baseline_}
 		\vspace{-1em}
 		\begin{tabular}{l c  c | c  c c}
 			\toprule
 			\multirow{2}{*}{\textbf{Baselines}}&
 			\multicolumn{2}{c}{\textbf{Model Compression}}&\multicolumn{3}{c}{\textbf{MIA Defense}}\cr
 			\cmidrule(lr){2-3} \cmidrule(lr){4-6}
 			& \textit{Pruning} & \textit{KD} &
 			\textit{DP} & \textit{AdvReg} & \textit{DMP} \\
 			\midrule
 			\textit{Pr-DP} & \Checkmark & & \Checkmark &  & \\
 			\textit{Pr-AdvReg} & \Checkmark &  & & \Checkmark &\\
 			\textit{Pr-DMP} & \Checkmark &  &  &  & \Checkmark\\ 
 			\textit{KD-DP} & &\Checkmark & \Checkmark&  & \\ 
 			\textit{KD-AdvReg} &  & \Checkmark & & \Checkmark  & \\ 
 			\textit{KD-DMP} & & \Checkmark  &  &  &\Checkmark \\ 
 			\bottomrule
 		\end{tabular}
 	\end{table}

 	\subsection{Baselines} \label{baseline}
 	
 	While few existing studies focus on safe model compression, we thus formulate several \textit{two-step} baseline methods by combining state-of-the-art model compression and MIA defense techniques.
 	
 	In the first step, we need to choose one \textbf{compression} method $\mathcal{C}$. We consider two widely used and effective choices to compress big models, \ie, \textit{pruning}~\cite{pruning_han2015deep} and \textit{knowledge distillation (KD)}~\cite{KD_hinton2015distilling}.  For pruning, We adopt the `pretrain$\to$prune$\to$fine-tune' paradigm. Firstly, we pretrain a big model from scratch. Then we leverage magnitude-based pruning and compress this full model to a certain sparsity. Then we fine-tune the pruned model to recover model performance. 
 	 For KD,  we design a small dense model as the student model given the sparsity requirement. Then, we use the well-trained big model as the teacher model to help the student model's training.
 	
 	In the second step, we should select one MIA \textbf{defense} method $\mathcal{D}$. We test three defense algorithms including \textit{differential privacy (DP)}~\cite{dp_abadi2016}, \textit{adversary regularization (AdvReg)}~\cite{ResAdv}, and  \textit{distillation for membership privacy (DMP)}~\cite{kd_membership}. DP adds noise to gradients while training. AdvRes adds membership inference gain $G_A$ of the attacker to the loss function of the target model as a regularization term, thereby forming a min-max game. 
 	DMP first trains an unprotected (vanilla) teacher model. Then, it utilizes extra unlabeled data and adopts knowledge distillation to force the student \ie, the target model, to simulate the output of this teacher. 
 	
 	Finally, we group two-step techniques by enhancing the training or fine-tuning process in $\mathcal{C}$ with the defense technique in $\mathcal{D}$. For example, the \textit{Pr-DP} baseline adopts the `pretrain$\to$prune$\to$DP-based fine-tune' process; \textit{KD-AdvReg} trains the student compressed model with the help of the big teacher model in a min-max game manner. All the six two-step baselines are listed in Table~\ref{baseline_}.

 	\textbf{Baseline 7 (MIA-Pr~\cite{Pruning_IJCAI}).}$\:$ This baseline is originally proposed to defend MIA, and its main technique is pruning. Hence, although the paper~\cite{Pruning_IJCAI} does not specify the usage for model compression, it can naturally reduce the model size. We use this method as a baseline to prune the big model until satisfying the sparsity requirement. Note that when pruning the big model, \textit{MIA-Pr} optimizes the MIA defense effect, while the two-step baseline \textit{Pr-X} (the first step is pruning) optimizes the task performance (\eg, classification accuracy).

\subsection{Implementation Details}
We perform all the experiments using Pytorch 1.8. on Ubuntu 20.04. For CV tasks, we use NVIDIA  1080Ti with CPU of Intel Xeon Gold 5118 (4 cores, 2.3GHz) and 16GB memory. We train big models 
with batch size $128$ for $200$ epochs. For NLP tasks, we use NVIDIA 3090 with CPU of Intel Xeon Gold 5218 (6 cores, 2.3GHz) and  40 GB memory. The batch size is 256 and the training lasts for 10 epochs. 

In \textit{MIA-SafeCompress}, we adopt SGD as the optimizer to train the target sparse model. We set the learning rate to 0.1 and adjust it via a multi-step decay strategy. 
To simulate the attacker, following previous work~\cite{ResAdv}, we adopt ReLu as the activation function in our attack neural network model. All the network weights are initialized with normal distribution with mean 0 and standard deviation 0.01, and all biases are set to 0 by default. The batch size is 128. We use the Adam optimizer with the learning rate of 0.001 all the time and we train the attack model 100 epochs. During the training process, we ensure that every training batch contains the same number of member and non-member data samples, aiming to prevent the attack model from being biased toward either side.
	\section{Experiments}
	\begin{figure}
		\centering{\includegraphics[width=1\linewidth]{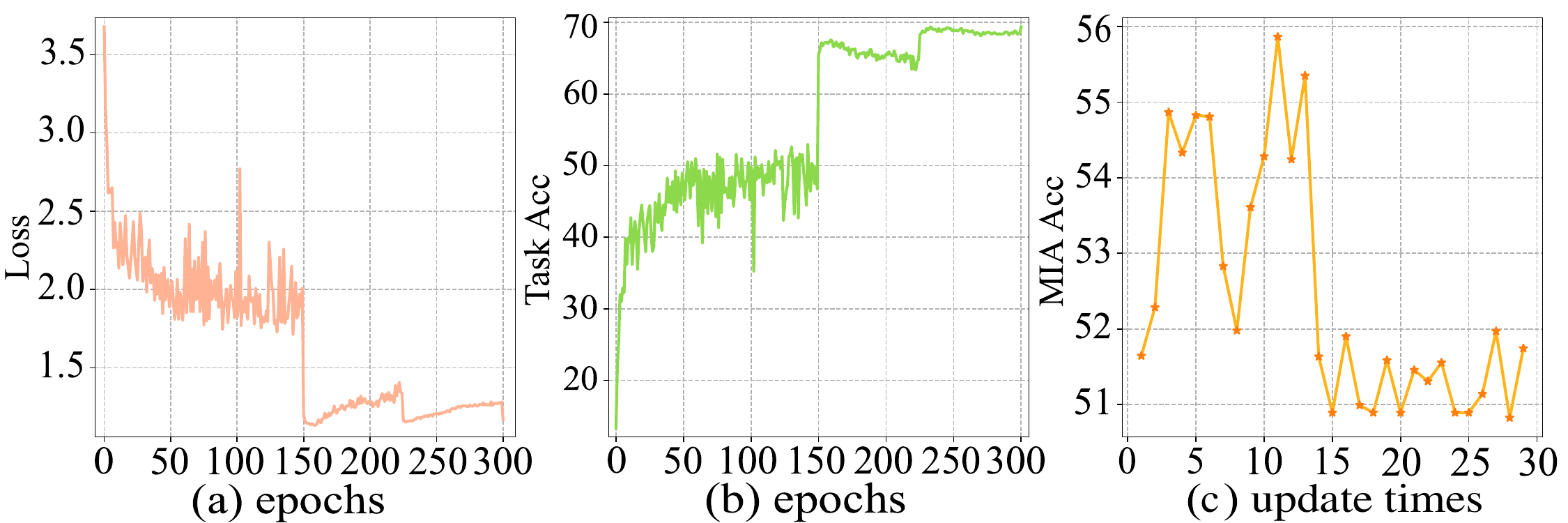}}
		\vspace{-2em}
		\caption{Visualization of the learning procedure.}
		\label{visualization}
		\vspace{-1em}
	\end{figure}
	\subsection{Learning Procedure Visualization}
	To better understand the learning  procedure of \textit{MIA-SafeCompress}, we train VGG16 on CIFAR100 with sparsity set to 0.05, and depict how loss, Task Acc, and MIA Acc vary when the model training and updating continues, as shown in Figure~\ref{visualization}. For each sparse model, we train about 10 epochs (\ie, about 4,000 iterations with batch size 128) and then update its sparse structure according to safety tests. That is, 10-epoch training usually triggers one update.
	We can observe that the model gradually converges as training progresses. 
	Specifically, by following the DST manner, \textit{MIA-SafeCompress} enables the target model to explore as many candidate sparse structures as possible.
	Then, with the help of safety tests, the sparse model structure can generally evolve into better ones (with higher Task Acc and lower MIA Acc) step by step. 

	\subsection{Regularization Term Selection}
	To decide the regularization term adopted in \textit{MIA-SafeCompress}, we conduct the experiments
	on CIFAR100. The results are reported in Table~\ref{regularization}. We also compare our method with widely-used L2 regularization.  Results show that our proposed \textit{re2} regularization works the best. Therefore, we choose \textit{re2} as an optional regularization term and report its results in our following experiments.
	\setlength{\tabcolsep}{0.15cm}\begin{table}[t]
		\caption{Comparison with different regularization methods. The best results are marked in \underline{\textbf{bold}}. 
		}
		\label{regularization}
		\vspace{-1em}
		\begin{tabular}{l  l  l l l }
			\toprule
			\textit{Regularization} & 
			\textit{no} &
			\textit{re1} &
			 \textbf{\textit{re2}} &
			\textit{L2} \\
			\midrule
			\textit{Task Acc} & 69.52\% & 69.89\% & $\underline{\textbf{69.91\%}}$ 
			& 68.86 \% \\
			\textit{MIA Acc} & 51.75\% & 51.94\% & $\underline{\textbf{51.54\%}}$
			 & 53.33\% \\
			 \midrule
			 \textit{TM-score} & 1.34 & 1.35 &  $\underline{\textbf{1.35}}$			 & 1.29 \\
			\bottomrule
		\end{tabular}
	\end{table}

	\subsection{Main Results on CV Datasets} \label{comparison_baselines_cv}
	
	The results of three CV datasets are illustrated as follows.
	
	\subsubsection{CIFAR100 (VGG16)}  
\begin{figure}[t]
	\centering{\includegraphics[width=1\linewidth]{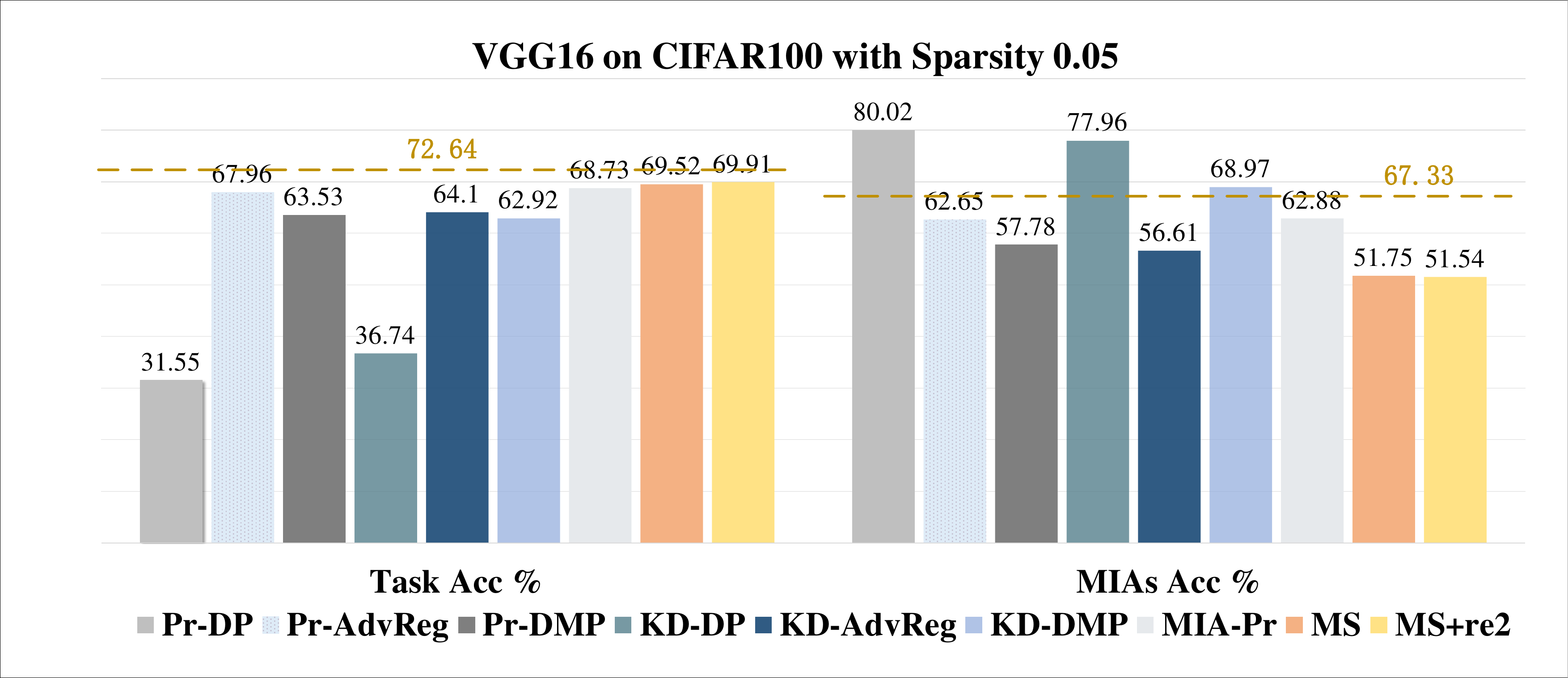}}
	\vspace{-2em}
	\caption{Results on CIFAR100 with sparsity 0.05 (MS: MIA-SafeCompress, dash line: uncompressed VGG16).}
	\label{vgg_0.05}
	\vspace{-.5em}
\end{figure}

Firstly, we report the Task Acc and MIA Acc on CIFAR100 with sparsity 0.05 using VGG16. As illustrated in Figure~\ref{vgg_0.05}, the results indicate that \textit{MIA-SafeCompress} outperforms seven baselines in Task Acc while alleviating MIA risks remarkably. Specifically, \textit{MIA-SafeCompress} produces 51.75\% for MIA Acc (just a bit higher than random guess), decreasing by 28.27\%, 10.9\%, 6.03\%, 26.21\%, 4.86\%, 17.22\%, and 11.13\% compared with seven baselines, respectively. Moreover, equipping \textit{MIA-SafeCompress} with \textit{re2} leads to an additional reduction of 0.21\% in MIA Acc while Task Acc rises 0.39\%. Interestingly, when compared to the uncompressed VGG16, \textit{SafeCompress} reduces MIA risks to a large extent while sacrificing Task Acc only a little. This highlights the practicality of \textit{SafeCompress} in generating a small and safe model with competitive task performance.  
	
\begin{figure*}
		\centering{\includegraphics[width=1\linewidth]{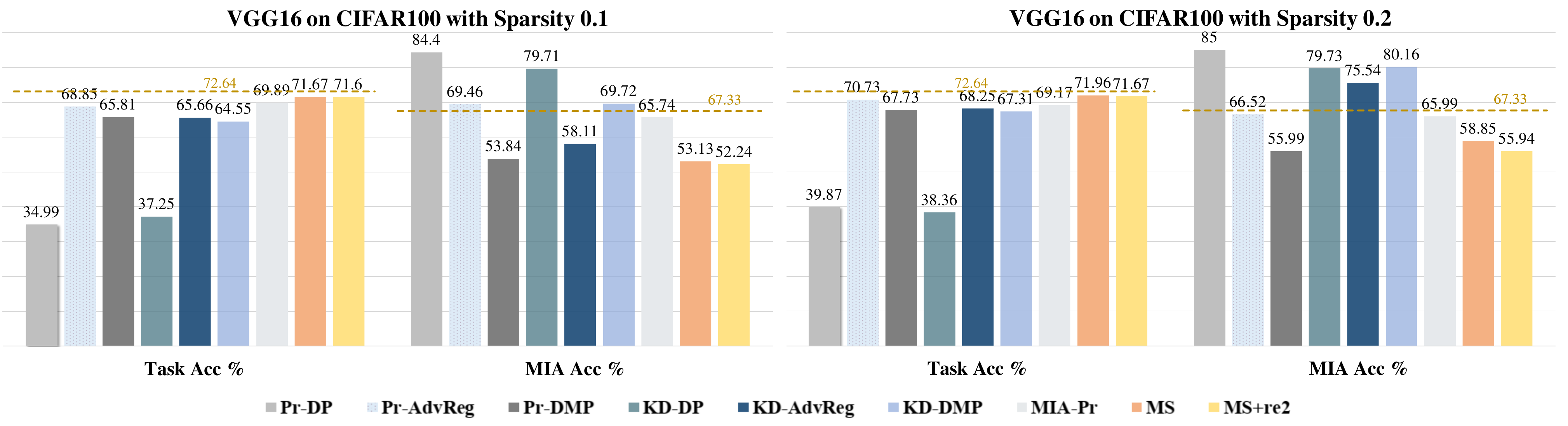}}
		\vspace{-2em}
		\caption{Results on CIFAR100 with sparsity 0.1 and 0.2 (MS: MIA-SafeCompress, dash line: uncompressed VGG16).}
		\label{vgg_0.1_0.2}
		\vspace{-1em}
	\end{figure*}

	Further, to validate the effectiveness on different sparsity requirements, we report the results in Figure~\ref{vgg_0.1_0.2} with sparsity 0.1 and 0.2. Consistent with the results of sparsity 0.05, \textit{MIA-SafeCompress} still beats all the baselines by achieving higher Task Acc and lower MIA Acc.
	We also observe that when sparsity increases, Task Acc of \textit{MIA-SafeCompress} increases from 69.52\% (sparsity 0.05) to 71.96\% (sparsity 0.2); meanwhile, MIA Acc also goes up from 51.75\% (sparsity 0.05) to 58.85\% (sparsity 0.2), indicating that the compressed model becomes more vulnerable with increasing sparsity. Such a phenomenon may be incurred by the extra information (\ie, more parameters) kept in the compressed model when sparsity rises --- some of the 
	extra information may be generalizable so that task performance is enhanced; other information might be specific to training data, thus leading to higher MIA risks. 
	
	Finally, to overview the performance-safety trade-off in one shot, we report TM-score (Task Acc divided by MIA Acc, Eq.~\eqref{score}) in Table~\ref{vgg_tm-score}. Clearly, \textit{MIA-SafeCompress (re2)} outperforms all the baselines significantly and consistently under three sparsity settings by making a better trade-off between performance and safety. 
	
	\setlength{\tabcolsep}{0.12cm}\begin{table}
	\small
		\caption{TM-score on CIFAR100. The best results are marked in \underline{\textbf{bold}}.}
		\label{vgg_tm-score}
		\vspace{-1em}
		\begin{tabular}{l  c  c  c }
			\toprule
			\textbf{Method}  & \textbf{Sparsity=0.05} & \textbf{Sparsity=0.1} & \textbf{Sparsity=0.2} \\
			\midrule
			\textit{VGG16} (uncompressed) & 1.08 & 1.08 & 1.08  \\
			\midrule
			\textit{Pr-DP} & 0.39 & 0.41 & 0.47  \\
			\textit{Pr-AdvReg} & 1.08
			 & 0.99 & 1.06 \\
			\textit{Pr-DMP} & 1.10
			 & 1.22 & 1.21 \\
			\textit{KD-DP} & 0.47 & 0.47 & 0.48 \\
			\textit{KD-AdvReg} & 1.13 & 1.13 & 0.90 \\
			\textit{KD-DMP} & 0.91 & 0.93 & 0.84 \\
			\textit{MIA-Pr} & 1.09 & 1.06 & 1.05 \\
			\midrule
			\textbf{\textit{MIA-SafeCompress}} & 1.34 & 1.35 & 1.22\\
			\textbf{\textit{+ re2}} & $\underline{\textbf{1.35}}$ &$\underline{\textbf{1.37}}$  & $\underline{\textbf{1.28}}$  \\
			\bottomrule
		\end{tabular}
		\vspace{-1em}
	\end{table}

	\subsubsection{CIFAR10 (AlexNet)} 
	The results on CIFAR10 are presented in Table~\ref{alexnet_0.05_0.1}. It can be seen  that \textit{MIA-SafeCompress} obtains almost the best performance in Task Acc and maintains a pretty strong defensive ability when sparsity is 0.05. Thanks to the excellent effectiveness in both aspects, \textit{MIA-SafeCompress} produces the highest TM-score, showing its outstanding ability to make the performance-safety trade-off. When the sparsity is set to 0.1, our method produces 85.31\% for Task ACC, slightly inferior to the best result (86.84\%) produced by KD-AdvReg. However, \textit{MIA-SafeCompress} decreases MIA Acc to 52.37\%, much lower than KD-AdvReg (55.59\%), leading to the highest TM-score again.
	\setlength{\tabcolsep}{0.3cm}\begin{table*}
		\caption{Task Acc (performance), MIA Acc (safety) and TM-score results on CIFAR10. The best results are marked in \underline{\textbf{bold}}.}
		\label{alexnet_0.05_0.1}
		\vspace{-1em}
		\begin{tabular}{l  c  c  c c c  c  c}
			\toprule
			\multirow{2}{*}{\textbf{Method}}&
			\multicolumn{3}{c}{\textbf{Sparsity=0.05} }& &\multicolumn{3}{c}{\textbf{Sparsity=0.1}}\cr
			\cmidrule(lr){2-4} \cmidrule(lr){6-8}
			& \textit{Task Acc} & \textit{MIA Acc} & \textit{TM-score} &
			& \textit{Task Acc} & \textit{MIA Acc} & \textit{TM-score} \\
			\midrule
			\textit{AlexNet} (uncompressed) & 87.41\% & 58.34\% & 1.50 & & 87.41\% & 58.34\% & 1.50 \\
			\midrule
			\textit{Pr-DP} & 54.57\% & 68.35\% & 0.80 & & 62.87\% &66.49\% & 0.95  \\
			\textit{Pr-AdvReg} & 82.72\% & $\underline{\textbf{52.74\%}}$ & 1.57& & 86.19\%  & 54.83\% & 1.57  \\
			\textit{Pr-DMP} & 81.66\% & 55.63\% & 1.47 & & 83.73\% & 54.05\%& 1.55  \\
			\textit{KD-DP} & 72.09\% & 59.94\% & 1.20 & & 73.27\% & 60.42\% &  1.21 \\
			\textit{KD-AdvReg} & 80.47\% & 54.38\% & 1.48 & & $\underline{\textbf{86.84\%}}$ & 55.59\% & 1.56  \\
			\textit{KD-DMP} & 82.23\% & 61.63\% & 1.33 & & 86.26\% & 63.45\% & 1.36  \\
			\textit{MIA-Pr} & 82.26\% & 53.15\% & 1.55 & & 86.11\%& 56.12\% &  1.53 \\
			\midrule
			\textbf{\textit{MIA-SafeCompress}} & 83.94\% & 52.97\% &  $\underline{\textbf{1.59}}$ & & 85.31\% & $\underline{\textbf{52.37\%}}$  &$\underline{\textbf{1.63}}$  \\
			\textbf{\textit{+ re2}} & $\underline{\textbf{84.00\%}}$ & 53.12\% & 1.58 & &85.38\% & 53.17\% & 1.61 \\
			\bottomrule
		\end{tabular}
	\end{table*}

	\subsubsection{Tinyimagenet (ResNet18)} 
	We present the results of Tinyimagenet in Table~\ref{resnet_0.05_0.1}. Excitingly, \textit{MIA-SafeCompress} produces the best accuracy for Task Acc in both sparsity 0.05 and 0.1, outperforming all the baselines by a large margin. Also, \textit{MIA-SafeCompress} obtains competitive MIA Acc in both sparsity settings compared to the best defense effects of baselines. Specifically, \textit{MIA-SafeCompress} achieves 52.46\% (sparsity 0.05) and 52.79\% (sparsity 0.1) in MIA Acc; The best MIA Acc results among baselines are 52.32\% (sparsity 0.05) and 52.44\% (sparsity 0.1). The gaps are tiny (0.14\% and 0.35\%). When employing \textit{re2}, the MIA Acc of \textit{MIA-SafeCompress} declines further, leading to the lowest MIA Acc. Then, \textit{MIA-SafeCompress (re2)} achieves the best TM-score in both sparsity settings.
	\setlength{\tabcolsep}{0.3cm}\begin{table*}
		\caption{Task Acc (performance), MIA Acc (safety) and TM-score results on Tinyimagenet. The best results are marked in \underline{\textbf{bold}}.}
		\vspace{-1em}
		\label{resnet_0.05_0.1}
		\begin{tabular}{l  c  c  c c c  c  c}
			\toprule
			\multirow{2}{*}{\textbf{Method}}&
			\multicolumn{3}{c}{\textbf{Sparsity=0.05} }& &\multicolumn{3}{c}{\textbf{Sparsity=0.1}}\cr
			\cmidrule(lr){2-4} \cmidrule(lr){6-8}
			& \textit{Task Acc} & \textit{MIA Acc} & \textit{TM-score} &
			& \textit{Task Acc} & \textit{MIA Acc} & \textit{TM-score} \\
			\midrule
			\textit{ResNet18} (uncompressed) & 65.48\% & 69.73\% & 0.94 & &  65.48\% & 69.73\% & 0.94 \\
			\midrule
			\textit{Pr-DP} & 19.26\% & 71.07\% & 0.27 & & 24.56\% &74.17\% & 0.33  \\
			\textit{Pr-AdvReg} & 60.10\% & 57.62\% & 1.04 & & 61.16\%  & 63.93\% & 0.96  \\
			\textit{Pr-DMP} & 55.56\% & 60.03\% & 0.92 & & 59.61\% & 64.56\%& 0.92  \\
			\textit{KD-DP} & 17.10\% & 52.32\% & 0.33 & & 17.45\% & 52.44\% &  0.33 \\
			\textit{KD-AdvReg} & 52.34\% & 53.27\% & 0.98 & & 54.48\% & 53.60\% & 1.02  \\
			\textit{KD-DMP} & 53.71\% & 57.18\% & 0.94 & & 57.16\% & 55.65\% & 1.03  \\
			\textit{MIA-Pr} & 58.36\% & 57.92\% & 1.01 & & 60.91\%& 61.03\% &  1.00 \\
			\midrule
			\textbf{\textit{MIA-SafeCompress}} & $\underline{\textbf{63.81\%}}$ & 52.46\% &  1.22 & & $\underline{\textbf{65.15\%}}$ & 52.79\% &1.24  \\
			\textbf{\textit{+ re2}} & 63.45\% &  $\underline{\textbf{51.27\%}}$  &$\underline{\textbf{1.24}}$ & &$\underline{\textbf{65.15\%}}$  & $\underline{\textbf{52.32\%}}$ & $\underline{\textbf{1.25}}$ \\
			\bottomrule
		\end{tabular}
	\end{table*}

    \subsection{Time Consumption}
	Table~\ref{time} reports the time consumption for each method to get a compressed model (sparsity = 0.05) on CIFAR100. In general, almost all the methods (except \textit{Pr-DP}) can finish compressing in 1--5 hours, and the time consumption of \textit{MIA-SafeCompress} is comparable to that of others. 
	Considering that \textit{MIA-SafeCompress} can achieve the best performance-safety balance (from our previous experiment results), the time consumption of \textit{MIA-SafeCompress} is generally acceptable in practice. Note that we only need to run \textit{MIA-SafeCompress} once to get the compressed model, and the model can then be deployed in hundreds of thousands of devices repeatedly. 
	\setlength{\tabcolsep}{0.2cm}\begin{table*}[h]
	\small
		\caption{Time consumption on CIFAR100. VGG16 is the uncompressed model. Others are compressed with sparsity 0.05.}
		\vspace{-1em}
		\label{time}
		\begin{tabular}{c  c  c c  c  c c c c c}
			\toprule
			\textit{Method} & 
			\textit{VGG16} (uncompressed)  &
			\textit{Pr-DP} &
			\textit{Pr-AdvReg} &
			\textit{Pr-DMP} &
			\textit{KD-DP} &
			\textit{KD-AdvReg} &
			\textit{KD-DMP} &
			\textit{MIA-Pr} &
			\textit{\textbf{MIA-SafeCompress (re2)}}
			\\
			\midrule
			\textit{Time} (h) & 1.05   & 16.90  & 2.92
			& 4.75 & 4.89
		
			& 2.22 & 1.42 & 2.37 & 3.98 \\
			\bottomrule
		\end{tabular}
	\end{table*}

	\subsection{Flexibility with Other Training Tricks}
	It is worth noting that \textit{SafeCompress} is also flexible to incorporate other training tricks not mentioned in previous experiments. To illustrate this flexibility, we try to incorporate five other widely-used training tricks in \textit{MIA-SafeCompress}, including \textit{dropout}~\cite{srivastava2014dropout}, \textit{data augmentation} (\textit{Aug}, including random cropping, resizing, and flipping), \textit{adversary regularization (AdvReg)}, \textit{sparse model initialization with the big model's parameter values (BigPara)}, and \textit{knowledge distillation (KD)} where the sparse model's prediction score approximates the big model's~\cite{KD_hinton2015distilling}. Table~\ref{flexsibility} shows the results of CIFAR100. 
	
	Compared to our \textit{MIA-SafeCompress} implementation in main experiments (\ie, \textit{re2}), we find that most training tricks achieve similar TM-scores. This indicates that \textit{SafeCompress} is compatible with most training tricks. Then, in practice, given a specific dataset/task, we could enumerate different combinations of training tricks and find the best one in the \textit{SafeCompress} framework. 
	
	Another interesting observation is that \textit{data augmentation significantly increases MIA Acc} (\ie, reduces safety) for the final compressed model. The possible reason is that data augmentation allows training samples to be memorized more easily, as the variants of original samples (\eg, flipping and resizing) are also used for training. Notably, while not focusing on compressed sparse models, a recent paper~\cite{yu2021does} also points out that data augmentation may lead to a significantly higher MIA risk. Inspired by \cite{yu2021does}, \textit{SafeCompress} may also be extended to a useful framework to benchmark different training tricks' impacts on compressed models' safety, which would be an interesting future direction.
	\setlength{\tabcolsep}{0.1cm}\begin{table}[t]
	    \small
		\caption{Using other training tricks in \textit{MIA-SafeCompress}.}
		\vspace{-1em}
		\label{flexsibility}
		\begin{tabular}{l l l l l l l l}
			\toprule
			\textit{Trick} & 
			 \textit{Dropout} &
			  \textit{Aug} &
			  \textit{AdvReg} &
			\textit{BigPara} &
			  \textit{KD}&
			\textit{\textbf{re2}} 
			\\
			\midrule
			\textit{Task Acc}  & 
			69.43\%
			& 62.03\% & 68.86\% & 68.52\% & $\underline{\textbf{70.02\%}}$ & 69.91\%  \\
			\textit{MIA Acc}   &
			53.94\% &
			65.46\% & 52.96\% & 52.05\% & 53.10\% &  $\underline{\textbf{51.54\%}}$\\
			\midrule
			\textit{TM-score} &
			1.29 &
			0.95 & 1.30& 1.32 & 1.32 & $\underline{\textbf{1.35}}$  \\
			\bottomrule
		\end{tabular}
		\vspace{-2em}
	\end{table}
	 
	
	\subsection{Results on NLP Datasets}
	To further validate the effectiveness and generalization of \textit{MIA-SafeCompress}, we conduct experiments on two NLP datasets.
	
	\subsubsection{Yelp-5 (BERT)} The results are reported in Table~\ref{bert_0.5}. It can be seen that \textit{MIA-SafeCompress} produces a competitive Task Acc. At the same time, it decreases MIA Acc by 6.78\% compared to the uncompressed model, outperforming all the baselines. When attaching \textit{re2} to \textit{MIA-SafeCompress}, we get the lowest MIA Acc. Besides, \textit{MIA-SafeCompress} achieves the highest TM-score, indicating its ability to balance task performance and safety.

	\subsubsection{AG-News (Roberta)} As indicated in Table~\ref{roberta_0.5}, \textit{MIA-SafeCompress} achieves 87.5\% for Task Acc, slightly inferior to the highest 88.10\% (KD-AdvReg). However, our method decreases MIA Acc to 55.28\%, much lower than KD-AdvReg (56.56\%). Moreover, \textit{MIA-SafeCompress} outperforms all the baselines in MIA Acc, thus resulting in a high TM-score (1.58). Moreover, incorporating \textit{MIA-SafeCompress} with \textit{re2} leads to a further enhancement in all the metrics.

	\section{Related Work}
	\subsection{Membership Inference Attacks and Defenses}
	Membership inference attacks (MIAs)~\cite{Shaow_Learning}, which aim to infer whether a data record is used to train a model or not, have the potential to raise severe privacy risks to individuals. A prevalent attack fashion is to train a neural network via multiple shadow training~\cite{Shaow_Learning, ResAdv}, converting the task of recognizing member and non-member of training datasets to a binary classification problem. Unlike binary classifier-based MIA, another attack form, metric-based MIA that just computes metrics (\eg, prediction correctness~\cite{dp_yeom2018privacy} or confidence~\cite{pre_conf_ML_Leaks}, entropy~\cite{song2021systematic_entropy}, \etc) is simpler and less computational. However, the performance of metric-based attacks is inferior to classifier-based attacks.  Recently, as more and more efforts have been devoted to the field, various defense methods have been presented. For example, differential privacy~\cite{dp_abadi2016, dp_rahman2018} (known as DP) interrupts the attack model by adding noise to the learning object or output of the target model.  But the cost between utilization and defense is unacceptable~\cite{dp_jayaraman2019evaluating_unaccept}. Adversarial regularization~\cite{ResAdv}, known as \textit{AdvReg}, combines target model training process with attack model, formulating a min-max game. To improve the model utility,  Jia~\etal introduce Memguard~\cite{memguard_jia2019} which adds carefully computed noise to output of a model, aiming to defend the attack model while keeping performance. However, it is vulnerable to the threshold-based attack~\cite{song2021systematic_entropy}. More recently, model compression technologies (\eg, knowledge distillation~\cite{KD_hinton2015distilling} and pruning~\cite{pruning_han2015deep}) have been employed to protect member privacy. Based on knowledge distillation, Shejwalkar and Houmansadr~\cite{kd_membership} propose Distillation For Membership Privacy (DMP) defense method, where a teacher is trained to label an unlabeled reference dataset, and those with low prediction entropy are selected to train the target model. So it requires an extra unlabeled reference dataset. Further, Zheng~\etal propose complementary knowledge distillation (CKD) and pseudo complementary knowledge distillation (PCKD)~\cite{MIAs_KD_V2}, eliminating the need for additional public data by just transferring knowledge from a private training set. However, as knowledge distillation is an indirect learning strategy~\cite{Moonshine_NEURIPS2018_49b8b4f9}, some critical information may be lost during mimicking the teacher. Afterward, in~\cite{Pruning_IJCAI}, pruning is adopted to mitigate MIA while reducing model size simultaneously. But it mainly focuses on preventing membership inference attacks without explicitly considering the memory restriction. Besides, Yuan \& Zhang~\cite{pruning_defeat_yuan2022membership} show that pruning makes the divergence of prediction confidence and prediction sensitivity increase and vary widely among the different classes of member and no-member data. This characteristic may be manipulated by attackers. Our method is different from these works that mainly focus on defending against MIA. This research tries to address safe model compression problem. More specifically,  we aim to decrease the risk of privacy attack (\eg MIA) and  keep excellent performance when compressing a model. It is a bi-objective optimization problem.
	
	\setlength{\tabcolsep}{0.25cm}\begin{table}[t]
	\small
		\caption{Task Acc, MIA Acc, and TM-score on Yelp-5.}
		\label{bert_0.5}
		\vspace{-1em}
		\begin{tabular}{l  c  c  c}
			\toprule
			\multirow{2}{*}{\textbf{Method}}&
			\multicolumn{3}{c}{\textbf{Sparsity=0.5} }\cr
			\cmidrule(lr){2-4}
			& \textit{Task Acc} & \textit{MIA Acc} & \textit{TM-score}
		 \\
			\midrule
			\textit{BERT} (Uncompressed) & 62.21%
\% & 71.15\% & 0.87 \\
			\midrule
			\textit{Pr-DP} & 60.77\% & 64.79\% & 0.94  \\
			\textit{Pr-AdvReg} & 61.17\% & 73.34\% & 0.83\\
			\textit{Pr-DMP} & 61.32\% & 66.41\% & 0.92  \\
			\textit{KD-DP} & 58.21\% & 65.36\% & 0.89  \\
			\textit{KD-AdvReg} & 61.38\% & 67.40\% & 0.91 \\
			\textit{KD-DMP} & 59.69\% & 70.25\% & 0.85 \\
			\textit{MIA-Pr} & $\underline{\textbf{61.80\%}}$ & 73.85\% & 0.84 \\
			\midrule

			\textit{\textbf{MIA-SafeCompress}} & 61.66\% & 64.37\% &  $\underline{\textbf{0.96}}$ \\
			\textit{\textbf{+ re2}} & 61.61\% &  $\underline{\textbf{64.24\%}}$  &$\underline{\textbf{0.96}}$ \\
			\bottomrule
		\end{tabular}
		\vspace{-1.3em}
	\end{table}
	
	\subsection{Model Compression}
	Due to limited memory and computation resources,  
	model compression~\cite{pruning_han2015deep} plays a crucial role, especially when transformer-based big models~\cite{BERT_devlin2018bert,VIT_dosovitskiy2020vit,SWIN_liu2021swin} become the mainstream. To alleviate the issue, various methods have been proposed. For example, pruning~\cite{pruning_han2015deep,Slim_liu2017learning}, as a direct and effective method, removes unimportant weights or structures according to certain criteria (\eg, weight magnitude). Knowledge distillation~\cite{KD_hinton2015distilling,Tiny_BERT_jiao2019tinybert}, known as KD, transforms knowledge from a big model (we call it teacher) to a small model (we call it student) during training. Determining how to design a student model and distill knowledge are two essential questions to be answered while applying KD. Quantification~\cite{quantify_han2015learning,quantify_chen2015compressing} is much simpler, which converts a long storage width in memory to a shorter one. For example, it could convert float (64 bit) to an 8-bit integer~\cite{8_quantization_jacob2018quantization}, even a binary (1 bit)~\cite{binary_hubara2016binarized,tenary_li2016ternary}. In addition, dynamic sparse training (denoted as DST), as a new compression method, is first proposed in~\cite{DST_Begin_mocanu2018scalable} and achieves surprising performance, attracting much attention from researchers. Follow-up works further introduce weight redistribution~\cite{weight_redistribut_dettmers2019sparse, anvance_mostafa2019parameter}, gradient-based weight growth~\cite{weight_redistribut_dettmers2019sparse,evci2020rigging}, and extra weights update in the backward pass~\cite{backupdate_jayakumar2020top,backupdate_raihan2020sparse} to improve the sparse training performance. More recently, by developing an independent framework, a truly sparse neural network without masks with over one million neurons can be trained on a typical laptop~\cite{liu2021sparse}. Hence, the DST paradigm has shown great potential to deploy neural network models into edge devices at a large scale. However, while there exist numerous works in model compression that balance size and performance, privacy safety is not well considered. Differing from these works, our work considers both safety and performance during model compression.
	
		\setlength{\tabcolsep}{0.2cm}\begin{table}[t]
		\small
		\caption{Task Acc, MIA Acc, and TM-score on AG-News.}
		\label{roberta_0.5}
		\vspace{-1em}
		\begin{tabular}{l  c  c  c}
			\toprule
			\multirow{2}{*}{\textbf{Method}}&
			\multicolumn{3}{c}{\textbf{Sparsity=0.5} }\cr
			\cmidrule(lr){2-4}
			& \textit{Task Acc} & \textit{MIA Acc} & \textit{TM-score}
			\\
			\midrule
			\textit{Roberta} (Uncompressed) & 89.20\% & 57.59\% & 1.55 \\
			\midrule
			\textit{Pr-DP} & 87.28\% & 56.07\% & 1.56 \\
			\textit{Pr-AdvReg} & 87.24\% & 57.58\% & 1.52 \\
			\textit{Pr-DMP} & 86.38\% & 56.31\% & 1.53 \\
			\textit{KD-DP} & 82.64\% & 56.13\% & 1.47 \\
			\textit{KD-AdvReg} & $\underline{\textbf{88.10}}$\% & 56.56\% & 1.56 \\
			\textit{KD-DMP} & 87.24\% & 56.64\% & 1.54 \\
			\textit{MIA-Pr} & 87.91\% & 57.04\% & 1.54 \\
			\midrule
			\textit{\textbf{MIA-SafeCompress}} & 87.50\% & 55.28\% &  1.58 \\
			\textit{\textbf{+ re2}} & 87.61\% &  $\underline{\textbf{54.79\%}}$  &$\underline{\textbf{1.60}}$ \\
			\bottomrule
		\end{tabular}
		\vspace{-1.3em}
	\end{table}
	
	
	\section{Limitations and Future Work}
	\label{sec:discussion}
	
	As one of the pioneering studies toward safe model compression, we have also identified several future work possibilities that may attract more effort to this direction.
	
	\textbf{Integrating heterogeneous attacks}. In practice, AI software would face various types of attacks. The \textit{SafeCompress} framework follows the \textit{attack configurability} principle to be able to be configured against different attacks, but it does not include an integration component to simultaneously consider heterogeneous attacks currently. A straightforward way is introducing multiple safety tests regarding heterogeneous attacks, and then using the weighted sum to combine these test measurements; the pitfall is the hardness of deciding the weights. More advanced strategies, such as bootstrapping according to different attacks' severity and commonality~\cite{han2019f}, may deserve future research.
	
	\textbf{Benchmarking performance-safety trade-offs between models}. In this work, for any dataset, we only use one state-of-the-art model as the original input big model. Prior research has pointed out that, when two models do the same task with similar performance, the model with more parameters may face higher MIA risks~\cite{nasr2019comprehensive}. Then, if we compress the two models with the same sparsity restrictions, would the statement still stand or be overturned? Analyzing this question would significantly help the model selection process in AI software deployment.

	
	

	
	

	\section{Conclusion}
	In this paper, we present a performance-safety co-optimization framework, called \textit{SafeCompress}, to address the \textit{safe model compression} problem, as it is critical for current large-scale AI software deployment. \textit{SafeCompress} is a test-driven sparse training framework, which can be easily configured to fight a pre-specified attack mechanism. Specifically, By simulating the attack mechanism, \textit{SafeCompress} performs both safety and performance tests, and iteratively updates the compressed sparse model. Based on \textit{SafeCompress}, we consider the membership inference attack (MIA), a representative attack form, and implement a concrete instance called \textit{MIA-SafeCompress} against MIA. Extensive experiments have been conducted using five datasets including three computer vision tasks and two natural language processing tasks. The results verify the effectiveness and generalization of our method. We also try to incorporate other training tricks into \textit{MIA-SafeCompress} and elaborate on how to adopt \textit{SafeCompress} to other attacks, showing the flexibility of \textit{SafeCompress}. As a pioneering study toward the safe model compression problem, we expect that our research can attract more effort to this promising direction in the new era when AI software is becoming more and more prevalent. 
	
	\section{Acknowledgement}
	We thank the anonymous reviewers for their constructive comments. We also thank Ruiqing Ding, Guanghong Fan, Duo Zhang, and Xusheng Zhang for useful discussions. This work is supported by the NSFC Grants no. 72071125 and 61972008.
	
	\section{Appendix}
	In this section, we try to configure \textit{SafeCompress} to another attack called white-box membership inference attack. We name this instance of \textit{SafeCompress} framework as \textit{WMIA-SafeCompress}. It is worth noting that white-box MIA is different from black-box attack. Specifically, the white-box setting allows an attacker to access to extra knowledge about the target model beyond outputs (\eg, hidden layer parameters~\cite{sablayrolles2019white} and gradient descents in training epochs~\cite{nasr2019comprehensive}).  Hence, it is regarded as a stronger attack manner. 
	
\begin{figure} 
		\centering{\includegraphics[width=.9\linewidth]{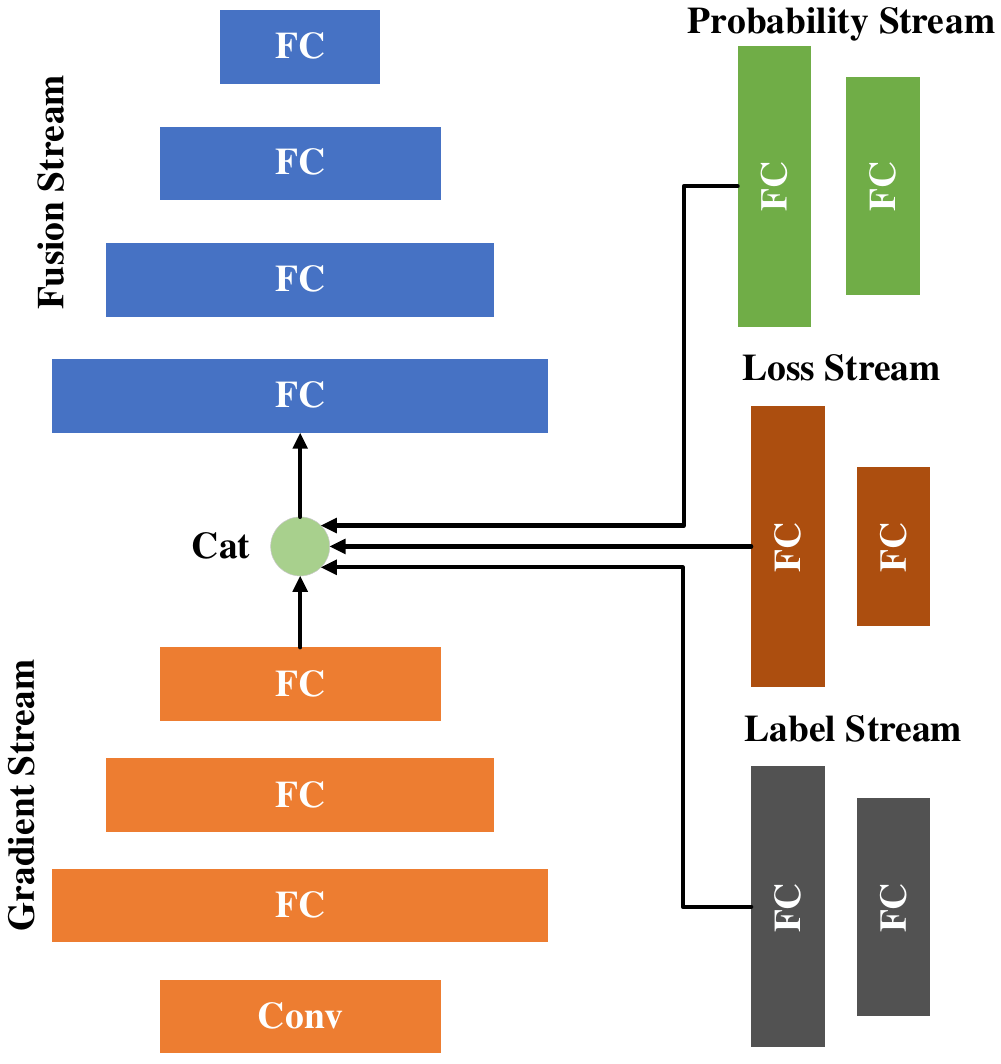}}
		\caption{The architecture of the simulated white-box MIA neural network model. 
		FC is fully connection layer. Conv means convolution layer.}
		\label{white_attack}
		\vspace{-1em}
	\end{figure}

	\textbf{Attacker Setting.} We follow the configuration instruction provided in section~\ref{sub:configure_other_attacks}. Concretely, we simulate the white-box attacker proposed in previous works~\cite{sablayrolles2019white, liu2021ml} while maintaining other settings in our paper. As illustrated in Figure~\ref{white_attack}, the simulated attacker contained five parts: probability stream, loss stream, gradient stream, label stream, and fusion stream. The first four streams are designed to process four different inputs respectively. They are the target sample’s ranked posteriors, loss value (\eg, classification loss), gradients of the parameters of the target model’s last layer, and one-hot encoding of its true label.  The fifth part, as the name suggests, fuses the extracted features from the first four streams and outputs a probability to indicate whether the sample is used in training or not. We use ReLU as the activation function for the attack model.
	
	\textbf{Results.} We choose the widely-used dataset CIFAR100 and the widely-concerned neural model VGG16 in this field to conduct the experiment with model sparsity set to 0.05. The results are reported in Table~\ref{white_vgg_0.05}. Our method produces 67.51\% for Task Acc, achieving a pretty competitive classification accuracy among all the baselines. In addition, \textit{WMIA-Safecompress} also decreases the MIA Acc by 15.36\% compared to the uncompressed VGG16. Although it is a little bit inferior (2.21\% lower) in MIA Acc than Pr-DMP,  our method maintains more performance (3.98\% higher) than Pr-DMP. Finally, we also calculate TM-score for each method to show its trade-off degree. It is observed that our method obtains the highest TM-score (1.20). Such results are reasonable as our framework, \textit{SafeCompress},  always targets bi-objective (safety and performance) optimization. 
Simultaneously, this experiment also indicates that our framework truly enables generalization to other, even stronger attacks. 

    \setlength{\tabcolsep}{0.2cm}\begin{table}[t]
		\small
		\caption{Task Acc, MIA Acc, and TM-score on VGG16.}
		\label{white_vgg_0.05}
		\begin{tabular}{l  c  c  c}
			\toprule
			\multirow{2}{*}{\textbf{Method}}&
			\multicolumn{3}{c}{\textbf{Sparsity=0.05} }\cr
			\cmidrule(lr){2-4}
			& \textit{Task Acc} & \textit{MIA Acc} & \textit{TM-score}
			\\
			\midrule
			\textit{VGG16} (Uncompressed) & 72.64\% & 71.58\% & 1.01 \\
			\midrule
			\textit{Pr-DP} & 31.55\% & 84.18\% & 0.37\\
			\textit{Pr-AdvReg} & 67.96\% & 67.05\% & 1.01\\
			\textit{Pr-DMP} & 63.53\% & $\underline{\textbf{54.01\%}}$ & 1.18 \\
			\textit{KD-DP} & 36.74\% & 77.40\% & 0.47 \\
			\textit{KD-AdvReg} & 64.10\% & 61.27\% & 1.05 \\
			\textit{KD-DMP} & 62.92\% & 63.30\% & 0.99 \\
			\textit{MIA-Pr} & $\underline{\textbf{68.73\%}}$ & 66.77\% & 1.03 \\
			\midrule
			\textit{\textbf{WMIA-SafeCompress}} &  67.51\% & 56.22\% & $\underline{\textbf{1.20}}$ \\
			\bottomrule
		\end{tabular}
	\end{table}

	\bibliographystyle{ACM-Reference-Format}
	\bibliography{bibfile}
\end{document}